\pdfoutput=1

\documentclass[11pt]{article}

\usepackage[]{ACL2023}
\usepackage{fancyhdr}

\usepackage{times}
\usepackage{latexsym}

\usepackage[T1]{fontenc}

\usepackage[utf8]{inputenc}

\usepackage{microtype}
\usepackage{inconsolata}
\usepackage{microtype}
\usepackage{booktabs}
\usepackage{graphicx}
\usepackage{bm}
\usepackage{algorithm}
\usepackage{algorithmic}
\usepackage{times}
\usepackage{latexsym}
\usepackage{amsmath,amssymb,amsfonts}
\usepackage{mathtools}
\usepackage{breqn}
\usepackage{subcaption}
\usepackage{multirow}
\usepackage{booktabs}
\usepackage{xcolor}
\usepackage[switch]{lineno} 
\usepackage{stmaryrd}

\usepackage{fancyhdr}

\fancypagestyle{firstpage}{%
  \fancyhf{} % Clear all headers/footers
  \fancyfoot[C]{In \textit{Findings of the Association for Computational Linguistics: IJCNLP-AACL 2023}, pages 161--170}
}
\title{Multilingual Non-Autoregressive Machine Translation \\ without Knowledge Distillation}

\author{
  \textbf{Chenyang Huang}\textsuperscript{*\rm 1},
  ~\textbf{Fei Huang}\textsuperscript{*\rm 4}, 
  ~\textbf{Zaixiang Zheng}\textsuperscript{\rm 3}, \\
  ~\textbf{Osmar R. Za\"iane}\textsuperscript{\rm 1}, 
  ~\textbf{Hao Zhou}\textsuperscript{\textdagger \rm 2},
  ~\textbf{Lili Mou}\textsuperscript{\rm 1} \\
    \textsuperscript{\rm 1}Dept. of Computing Science, Alberta Machine Intelligence Institute (Amii), University of Alberta \\
    \textsuperscript{\rm 2}Institute for AI Industry Research (AIR), Tsinghua University ~ \\
    \textsuperscript{\rm 3}ByteDance Research  ~
    \textsuperscript{\rm 4}Damo Academy, Alibaba \\
  \tt{chenyangh@ualberta.ca} ~ \tt{huangfei382@163.com} ~ \tt{zhengzaixiang@bytedance.com} \\
   \tt{zhouhao@air.tsinghua.edu.cn} ~ \tt{zaiane@ualberta.ca} ~ {\tt doublepower.mou@gmail.com} 
}

\makeatletter
\def\@maketitle{
  \newpage
  \null
  \vskip 2em%
  \begin{center}%
  \let \footnote \thanks
    {\Large \bfseries \@title \par}%
    \vskip 1.5em% 
    {\normalsize
      \lineskip .5em%
      \begin{tabular}[t]{c}%
        \@author
      \end{tabular}\par}%
    \vskip 1em%
  \end{center}%
  \par
  \vskip 1.5em% 
}
\makeatother

\begin{document}
\maketitle
\begingroup\def\thefootnote{*}\footnotetext{Work partially done during an internship at ByteDance.}\endgroup
\begingroup\def\thefootnote{\textdagger}\footnotetext{Work partially done while working at ByteDance.}\endgroup

\begin{abstract}
Multilingual neural machine translation (MNMT) aims at using one single model for multiple translation directions. Recent work applies non-autoregressive Transformers to improve the efficiency of MNMT, but requires expensive knowledge distillation (KD) processes. To this end, we propose an M-DAT approach to non-autoregressive multilingual machine translation. Our system leverages the recent advance of the directed acyclic Transformer (DAT), which does not require KD. We further propose a pivot back-translation (PivotBT) approach to improve the generalization to unseen translation directions. Experiments show that our M-DAT achieves state-of-the-art performance in non-autoregressive MNMT.\footnote{Our code and training/evaluation scripts are available at \url{https://github.com/MANGA-UOFA/M-DAT}}
\end{abstract}

\thispagestyle{firstpage}

\section{Introduction}
Multilingual neural machine translation (MNMT) aims at using a single model for multiple translation directions  \cite{firat-etal-2016-multi}. It has attracted the attention of the research community over the years \cite{adapter,zhang2021share}, and has been widely applied in the industry \cite{johnson-etal-2017-googles}.
Most state-of-the-art MNMT models are based on the autoregressive Transformer \cite[AT,][]{attentionisallyouneed}.
However, the inference of AT is slow, which results in significant latency in real-world applications \cite{gu2018nonautoregressive}. 

Recent work applies the non-autoegressive Transformer \cite[NAT,][]{gu2018nonautoregressive}, which generates target tokens in parallel 
for efficient inference. However, NAT often generates inconsistent sentences (e.g., with repetitive words). \newcite{qian-etal-2021-glancing} propose the Glancing Transformer (GLAT), which is trained in a curriculum learning fashion. It tackles the weakness of NAT by focusing less on the training samples that lead to inconsistent generalization.

To accelerate multilingual non-autoregressive translation, \newcite{song2022switchglat} propose a Switch-GLAT method, which is based on the Glancing Transformer, and is equipped with back-translation for data augmentation. 
To the best of our knowledge,
Switch-GLAT is currently the only non-autoregressive system for multilingual translation. 
However, it suffers from two drawbacks. First, Switch-GLAT requires sequence-level knowledge distillation \cite[KD,][]{kim-rush-2016-sequence} in every translation direction, which is inconvenient for multilingual tasks. Second, Switch-GLAT is unable to generalize to unseen translation directions (zero-shot translation), which is an essential aspect of multilingual machine translation systems \cite{johnson-etal-2017-googles,chen-etal-2017-teacher,gu-etal-2019-improved}.

In this work, we propose a multilingual Directed Acyclic Transformer (M-DAT) approach to non-autoregressive multilingual machine translation. Our system leverages the recent directed acyclic Transformer \cite[DAT,][]{dag}, which does not rely on KD. In addition, we propose a pivot back-translation (PivotBT) approach for the multilingual translation task. Specifically, we back-translate a target sentence to a randomly selected language to obtain an augmented source sentence. The newly generated source sentence and the original target sentence form a synthetic data sample. We observe that if the back-translation direction (e.g., German $\rightarrow$ Romanian) does not exist in the training set (i.e., zero-shot), the augmented source sentence will be of low quality. Therefore, our proposed PivotBT uses an intermediate language for the back translation (e.g., German $\rightarrow$ English $\rightarrow$ Romanian). Our PivotBT is efficient, as the inference of our non-autoregressive model is fast.

We evaluated M-DAT in both supervised and zero-shot translation directions.
In the supervised setting, our M-DAT achieves 0.4 higher BLEU scores than the previous state-of-the-art Switch-GLAT, while maintaining fast inference. Moreover, our M-DAT does not require KD, and is convenient to be trained on multilingual datasets.
In the zero-shot translation settings, our M-DAT is the first NAT model to effectively generalize to unseen translation directions, and even outperforms a strong autoregressive baseline, which is largely attributed to our proposed PivotBT.

\section{Related Work}

The non-autoregressive Transformer \cite[NAT,][]{gu2018nonautoregressive} predicts all target words in parallel to achieve fast inference, 
and has been applied to various text generation tasks, such as machine translation \cite{gu-kong-2021-fully,dslp}, summarization \cite{su-etal-2021-non,liu2022learning,liu2022character}, and dialogue generation \cite{zou-etal-2021-thinking,qi2021bang}.
However, the output quality of NAT models tends to be low \cite{stern,ghazvininejad-etal-2019-mask}, and as a remedy, the Glancing Transformerthe Glancing Transformer \cite[GLAT,][]{qian-etal-2021-glancing} applies an adaptive training algorithm that allows the model to progressively learn more difficult data samples.

Sequence-level knowledge distillation \cite[KD,][]{kim-rush-2016-sequence} is commonly used to improve the translation quality of non-autoregressive models.
As shown in \newcite{understandingKD}, KD data are less complex compared with the original training set, which is easier for NAT models. However, \newcite{lfw} find that the KD process tends to miss low-frequency words (e.g., proper nouns), which results in worse translation for NAT models. Therefore, there is a need to remove the KD process for NAT models.

The most related work to ours is Switch-GLAT \cite{song2022switchglat}. It combines the Glancing Transformer and knowledge distillation for multilingual machine translation tasks. In addition, Switch-GLAT is equipped with a back-translation technique to augment training data.

Our system is based on the directed acyclic Transformer \cite[DAT,][]{dag},
which expands the generation canvas to allow multiple plausible translation fragments. Then, DAT selects the fragments by predicting linkages, which eventually form an output sentence. 
% which consolidates multiple candidate target sequences into a directed acyclic graph, and linearizes the graph to be the output of the NAT. 
In this way, M-DAT is more capable of handling complex data samples, and does not rely on KD.

We propose PivotBT to augment the training data to improve the generalization of M-DAT. Our PivotBT is inspired by online back-translation \cite{zhang2020improving}, which extends the original back-translation \cite[BT,][]{bt} by randomly picking augmented directions. Different from the previous work, our approach applies pivot machine translation \cite{pivot} to improve the reliability of the back-translation directions that are unseen in the training set.

\begin{figure}[t]
  \centering
  \includegraphics[width=1.0\linewidth]{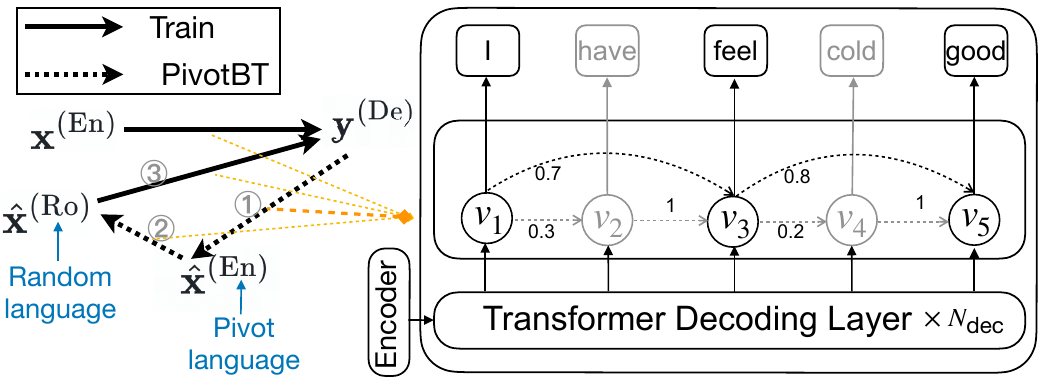}
  \vspace{-0.5cm}
  \caption{An example of our PivotBT augmenting a German sentence $\mathbf y$ to Romanian $\hat{\mathbf{x}}$, where English is used as the pivot language. The training and back-translation steps are accomplished by DAT \cite{dag}. 
  $N_{\text{dec}}$ is the number of decoding layers.}
  \label{fig:dat}
\end{figure}

\section{Methodology}

Multilingual translation handles multiple translation directions with a single model.
Suppose a data sample contains a source sentence $\mathbf x = (\mathrm x_1, \cdots, \mathrm x_{T_{\mathrm x}})$ and a target  sentence  $\mathbf y = (\mathrm y_1, \cdots, \mathrm y_{T_{\mathrm y}})$, where $T_{\mathrm x}$ and $T_{\mathrm y}$ denote the lengths. 
In addition, language tags $l_{\text{src}}$ and $l_{\text{tgt}}$ are given to indicate the languages of the source and target sentences, respectively.
A multilingual machine translation dataset $\mathcal{D}$ can be represented by
$\{(\mathbf x^{(i)}, l_{\text{src}}^{(i)}, \mathbf y^{(i)}, l_{\text{tgt}}^{(i)} ) \}_{i=1}^K$,
where $K$ is the size of the dataset.

Our multilingual Directed Acyclic Transformer (M-DAT) has an encoder--decoder architecture.
The encoder of M-DAT takes in an input sentence $\mathbf x^{(i)}$ and the target language tag $l_{\textrm{tgt}}^{(i)}$, whereas the decoder predicts the target-language words independently as the translation. 

\subsection{Directed Acyclic Transformer}
\label{sec:dat}
To train our system without KD, we adapt the recent directed acyclic Transformer \cite[DAT,][]{dag}, as it does not use KD but achieves comparable performance to autoregressive models on bilingual machine translation tasks (one direction per model).

In general, DAT expands its output canvas to generate multiple plausible translation fragments. Further, DAT predicts links to select the fragments, which form an output sentence. 
As seen in Figure~\ref{fig:dat}, DAT predicts extra words and forms the final generation ``I feel good'', using the predicted links.

Suppose the DAT decoder has $S$ generation steps ($S > T_{\mathrm y}$). For each step $s$ within  $ 1 \leq s \leq S$, DAT makes two predictions: word prediction $p_{\text{word}}^{(s)}(\cdot)$ and link prediction $p_{\text{link}}^{(s)}(\cdot)$.

The word prediction $p_{\text{word}}^{(s)}(\cdot)$ gives the distribution over possible words by mapping DAT's $s$th decoder state $\bm h_{s}$ to the probability distribution over the vocabulary, given by 
\begin{align}
  p_{\text{word}}^{(s)}(\cdot) = \operatorname{softmax}(\mathbf W_{\text{word}} \bm h_{s})
\end{align}
where $\mathbf W_{\text{word}}$ is a learnable matrix. 

The link prediction $p_{\text{link}}^{(s)}(\cdot)$ computes the distribution over the subsequent steps of the $s$th step, determining which follow-up step should be linked to the $s$th step.
Specifically, the link prediction leverages the attention mechanism \cite{BahdanauCB14}, which compares the $s$th step's hidden state $\bm h_{s}$ with the states of subsequent generation steps (from $s+1$ to $S$), given by 
\begin{align}
\begin{aligned}
  p_{\text{link}}^{(s)} (\cdot) = \operatorname{softmax}([\bm k_{s}^{\top} \bm q_{{s}+1};  \cdots; \bm k_{s}^{\top} \bm q_{{S}}])
\end{aligned} \label{eq:pos_prediction}
\end{align}
where $\bm k_{s} = \mathbf W_{\mathrm k} \bm h_{s}$ and $\bm q_{s} = \mathbf W_{\mathrm q} \bm h_{s}$. $\mathbf W_{\mathrm k}$ and $\mathbf W_{\mathrm q}$ are learnable matrices. The operation $[;]$ concatenates scalars into a column vector. 

Given a reference sequence $\mathbf y$ in the training set $\mathcal {D}$, DAT selects $T_{\mathrm y}$ of all $S$ generation steps to generate the words in $\mathbf y$, where the selected steps are connected by predicted links. We denote the indices of the selected steps by $\bm a = (a_1, \cdots, a_{T_{\mathrm y}})$, where $1 = a_1 < \cdots < a_{T_{\mathrm y}} = S$. We refer to each selection of the steps $\bm a$ as a \emph{path}.

Consider a groundtruth sequence $\mathbf y_{1:T_{\mathrm y}}$. The joint probability of the sequence, together with some path $\bm a_{1:T_{\mathrm y}}$, is
\begin{align}
\resizebox{.85\linewidth}{!}{$
  \begin{aligned}
    p(\mathbf y_{1:T_{\mathrm y}}, \bm a_{1:T_{\mathrm y}}) = \prod_{t=2}^{T_{\mathrm{y}}} p_{\text{link}}^{(a_{t-1})}(a_t) \prod_{t=1}^{T_{\mathrm y}} p_{\text{word}}^{(a_t)}(\mathrm y_t) 
  \end{aligned}
  $}
  \end{align}
where $p_{\text{link}}^{(a_{t-1})}(a_t)$ is the probability that the two generation steps $a_{t-1}$ and $a_t$ are linked up. Specially, $a_1$ is set to $1$, and is not considered as a random variable. $p_{\text{word}}^{(a_t)}(\mathrm y_t)$ is the probability of predicting the word $\mathrm y_t$ at the $a_t$th generation step.

Finally, the probability of generating the reference sentence $p(\mathbf y_{1:T_{\mathrm y}})$ is obtained by the marginalization of all possible paths, given by
\begin{align}
\resizebox{.75\linewidth}{!}{$
\begin{aligned}
    & p(\mathbf y_{1:T_{\mathrm{y}}} ) = \sum_{\boldsymbol a \in \Gamma_{S,T_{\mathrm y}} } p( \mathbf y_{1:T_{\mathrm y}}, \boldsymbol a_{1:T_{\mathrm y}}) \\
    & = \sum_{\boldsymbol a \in \Gamma_{S,T_{\mathrm y}} }  \prod_{t=2}^{T_{\mathrm{y}}} p_{\text{link}}^{(a_{t-1})}(a_t) \prod_{t=1}^{T_{\mathrm y}} p_{\text{word}}^{(a_t)}(\mathrm y_t)  \label{eq:dag}
\end{aligned}
$}
\end{align}
where $\Gamma_{S,T_{\mathrm{y}}} = \{ \bm a = ( a_1, \cdots, a_{T_{\mathrm y}}) | 1 = a_1 < \cdots < a_{T_{\mathrm{y}}} = S \}$ represents all paths of length $T_{\mathrm y}$.
The computation of (\ref{eq:dag}) is efficient through dynamic programming.\footnote{We refer readers to \newcite{dag}.}

Note that $p_{\text{word}}^{(s)}(\cdot)$ and $p_{\text{link}}^{(s)}(\cdot)$ are independently predicted for different generation steps; thus, DAT is non-autoregressive and is fast in inference.

\subsection{Pivot Back-Translation}
We propose a pivot back-translation (PivotBT) approach to improve the robustness of M-DAT. Following \newcite{zhang2020improving} and \newcite{song2022switchglat}, we augment the training data with back-translation \cite[BT,][]{bt}.
Specifically, a randomly selected language is chosen for such data augmentation. 

We observe that when the back-translation direction is unseen (i.e., zero-shot), the synthesized source sentence will be of low quality, which results in a less meaningful synthetic training sample. To this end, we propose to handle the zero-shot scenario by PivotBT, which uses an intermediate language as a pivot and performs multi-step back-translation.

Given a training sample $({\mathbf{x}}, l_{\text{src}}, {\mathbf{y}}, l_{\text{tgt}})$, we first randomly pick a language $l_{\text{aug}}$ from
the set of languages in the multilingual training set $\mathcal{D}$.
If the back-translation direction $ l_{\text{tgt}} \rightarrow  l_{\text{aug}}$ is in the training set, we directly back-translate ${\mathbf{y}}$ to ${\hat{\mathbf{x}}}$ of language $l_{\text{aug}}$.
Otherwise, we choose a pivot language $l_{\text{pivot}}$ (e.g., English) such that $ l_{\text{tgt}} \rightarrow  l_{\text{pivot}}$ and $ l_{\text{pivot}} \rightarrow  l_{\text{aug}}$ are both in the training set.\footnote{Most multilingual translation datasets are English-centric, where using English as $l_{\text{pivot}}$ guarantees the connection of $l_{\text{tgt}}$ and $l_{\text{aug}}$, which is the case of this work. In general, we can use multiple pivot languages to connect $l_{\text{tgt}}$ and $l_{\text{aug}}$ with multiple back-translation steps.} In this way, we are able to obtain an augmented source sentence $\hat{\mathbf{x}}$ by first translating ${\mathbf{y}}$ to ${\hat{\mathbf{x}}}_{\text{pivot}}$ of the intermediate language $l_{\text{pivot}}$, and then translating ${\hat{\mathbf{x}}}_{\text{pivot}}$ to the augmented source sentence ${\hat{\mathbf{x}}}$ of language $l_{\text{aug}}$. 
Finally, the newly synthesized sample $({\hat{\mathbf{x}}}, l_{\text{aug}}, {\mathbf{y}}, l_{\text{tgt}})$ is added to the training. 

In our PivotBT, the multi-step back-translation is conducted by M-DAT itself. Since M-DAT is fast in inference, the back-translation is also efficient.

We denote the loss of training the real samples in dataset $\mathcal D$ by $\mathcal L_{\text{real}}$ and that of the synthetic samples by $\mathcal L_{\text{PivotBT}}$. 
The overall training loss of our proposed system is $\mathcal L = \mathcal L_{\text{real}} + \lambda \mathcal L_{\text{PivotBT}}$, 
where $\lambda$ is a hyperparameter controlling the strength of the back-translation.

\begin{table}[t]
\centering
\resizebox{0.9\linewidth}{!}{
\begin{tabular}{llllr}  \toprule
\#    &     Model Variant   & EFD   & EFZ   & MANY  \\ \midrule
1    & M-AT  w/ standard layout       & \textbf{34.57} & \textbf{31.15} & \textbf{30.36} \\
2    & M-AT  w/ shallow decoder        &  34.19   & 30.87  & 29.20  \\
3 & Switch-GLAT$^*$    & 33.34 & 29.76 & 28.47 \\ \midrule
4 &  M-DAT w/ lookahead   & 33.72 & 30.39 & 28.69 \\
5 & ~~  w/ $n$-gram beam search & 33.83 & 30.55 & 29.73 \\ 
\bottomrule
\end{tabular}}
\vspace{-0.15cm}
\caption{BLEU scores on three WMT datasets.  $^*$Trained with sequence-level knowledge distillation.}
\label{tab:wmt_results_sum}
\end{table}

\section{Experiments}
\subsection{Setup}

We evaluated M-DAT on five datasets: WMT-EFD, WMT-EFZ, WMT-MANY, IWSLT, and Europarl. The three WMT corpora \cite{song2022switchglat} only contain supervised directions in the test set, whereas the test sets of IWSLT and Europarl \cite{disentangle-pos} contain both supervised directions and unseen directions (zero-shot). 
The training hyperparameters and evaluate metrics are strictly following \newcite{song2022switchglat} and \newcite{disentangle-pos}.
We provide more details in Appendices~\ref{app:datasets}~and~\ref{app:settings}.

\subsection{Main Results}
\textbf{Supervised Translation.}
Table~\ref{tab:wmt_results_sum} summarizes the BLEU scores of three WMT datasets.
We evaluated M-DAT with two decoding algorithms: \emph{lookahead} and \emph{$n$-gram beam search}. The lookahead method directly decodes the generated words in parallel, and jointly maximizes the probability of the next position and predicted tokens, whereas $n$-gram beam search generates a few candidate sentences and ranks them with an $n$-gram language model.
We observe that the generation quality with $n$-gram beam search is higher, which is consistent with \newcite{dag}.

We also see that M-DAT outperforms Switch-GLAT on average with both lookahead and beam search decoding methods.
This makes our M-DAT the state of the art in non-autoregressive multilingual translation. In addition, our system does not rely on KD, which makes it convenient to be trained on multilingual datasets. %  compared with Switch-GLAT. 

To compare M-DAT with the autoregressive multilingual Transformer \cite[M-AT,][]{johnson-etal-2017-googles}, we include two autoregressive Transformer-based variants: 1) the standard layout \cite{attentionisallyouneed}, which has the same number of encoder layers; and 2) the layout with a shallow decoder, which moves all but one decoding layers to the encoder. The layout with a shallow decoder is suggested by \newcite{Kasai0PCS21} as it achieves close results to the standard layout but is faster in inference. 
We observe that M-DAT is only slightly lower in BLEU scores compared with the M-AT models. 
This shows our M-DAT, in addition to its efficiency, largely closes the gap between AT and NAT in non-autoregressive multilingual translation tasks.

\begin{table}[t]
  \centering
  \resizebox{0.95\linewidth}{!}{
  \begin{tabular}{lrr|rr} \toprule 
  \multirow{2}{*}{\vspace{-0.2cm}Model Variant}     &   \multicolumn{2}{c}{IWSLT}    &    \multicolumn{2}{c}{Europarl} \\ \cmidrule{2-5}
                   & \!\!\!\!\!\!Supervised  & 0-Shot & Supervised  & 0-Shot  \\ \midrule
  M-AT w/ standard layout      & \bf{30.00}         & 12.87    &  \textbf{35.79}  & 15.84  \\  
  M-AT w/ shallow decoder      &   29.23    & 4.95   &  34.95    & 8.11 \\   
  Residual M-AT  & 29.72       & 17.67   &  35.18  & 26.13  \\ \midrule 
  M-DAT   w/ lookahead  & 28.58  & 18.53  &  34.83  &  25.86  \\ 
  ~~ w/ $n$-gram beam search   & 29.42   & \bf{19.35}  &  35.48 & \textbf{27.44} \\ \bottomrule
  \end{tabular}}
  \vspace{-0.15cm}
  \caption{BLEU scores on IWSLT and Europarl.}
  \label{tab:zero_shot}
\end{table}

\begin{table}[t]
\centering
\resizebox{0.85\linewidth}{!}{
\begin{tabular}{lrr} \toprule
Model Variant      & Latency (ms)  & Speedup \\ \midrule
M-AT  w/ standard layout         & 352.4     & 1.0$\times$   \\
M-AT   w/ shallow decoder    &  84.2     & 4.2$\times$   \\
Switch-GLAT        & \textbf{19.6}    & \textbf{18.0}$\times$    \\ \midrule
M-DAT w/ lookahead & 21.9    & 16.1$\times$    \\
~~ w/ $n$-gram beam search  & 67.6    & 5.2$\times$    \\  \bottomrule
\end{tabular}}
\vspace{-0.15cm}
\caption{Latency and speedup on WMT-EFD.}
\label{tab:speed}
\end{table}

\begin{figure}[t]
  \centering
    \includegraphics[width=0.75\linewidth]{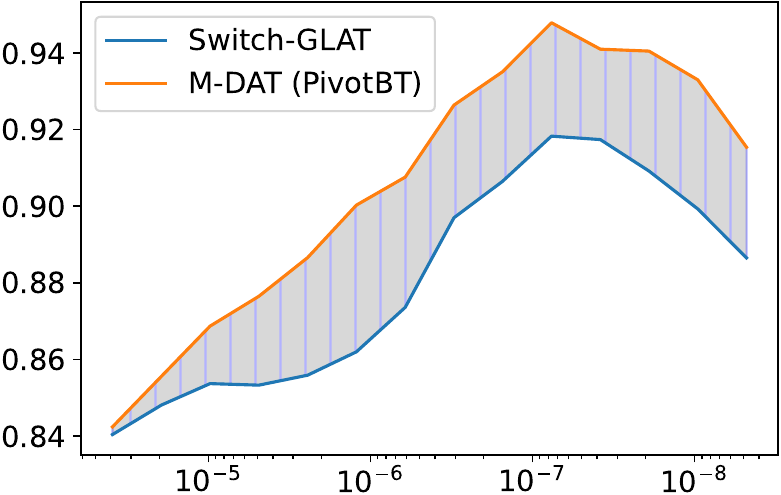}
    \vspace{-0.15cm}
    \caption{Comparison between M-DAT and Switch-GLAT in the preservation ratio of low-frequency words.}
    \label{fig:lfw}\vspace{-0.1cm}
\end{figure}

\textbf{Zero-Shot Translation.} The ability to generalize to unseen translation directions is important for multilingual models.  In this setting, we do not compare our model with Switch-GLAT as it fails to achieve reasonable performance.\footnote{As seen in the Table~7 of \newcite{song2022switchglat}, Switch-GLAT only obtained a 2.34 BLEU on a zero-shot dataset.} 
Instead, we compare M-DAT with the M-AT models \cite{johnson-etal-2017-googles}, and include a recent study, Residual M-AT \cite{disentangle-pos}, which relaxes the residual connections in the Transformer encoder to force the decoder to learn more generalized representations.\footnote{The numbers of Residual M-AT are based on our replication, and are close to those in \newcite{disentangle-pos}. }

As seen in  Table~\ref{tab:zero_shot},  the M-ATs are incapable of zero-shot translation, and are largely outperformed by the Residual M-AT. On the other hand, our M-DAT with the lookahead decoding outperforms Residual M-AT by 0.86 BLEU on the IWSLT dataset, although it is 0.27 lower on the Europarl dataset. With $n$-gram beam search, the improvement is 1.68 BLEU on IWSLT and 1.31 on Europarl. Our M-DAT is the first non-autoregressive model to outperform a strong AT baseline in zero-shot multilingual translation.\footnote{In our experiments, the autoregressive baseline models are not equipped with PivotBT, as their inefficient inference (as seen in Table~\ref{tab:speed}) makes the training with back-translation impractical. In fact, our PivotBT is another way to take advantage of the fast inference of non-autoregressive models.}

\textbf{Inference Speed.} We compare the inference speed on the test set of WMT-EFD and present the results in Table~\ref{tab:speed}. The batch size is set to 1 to mimic the real-world scenario, where the user requests come one after another \cite{gu2018nonautoregressive}. 
As seen, M-DAT with lookahead is about 16 times faster than the standard autoregressive baseline, and is about 4 times faster than  M-AT with a shallow decoder. Compared with Switch-GLAT,  M-DAT with lookahead is about the same efficiency. Admittedly, M-DAT with beam search is slower, but is still 5.2 times faster than the standard M-AT. In general, M-DAT obtains a good speed--quality trade-off.

\subsection{Analysis}

\textbf{Low-Frequency Words.} We analyze the generated text (on the WMT-EFD test set) of our M-DAT and Switch-GLAT. We followed \newcite{lfw}, and computed the percentage of a word being preserved in the translated sentence (in the corresponding language); then we grouped the words by their frequencies in the dataset.\footnote{We applied the  \texttt{FastAlign} alignment tool \cite{dyer-etal-2013-simple} to approximate word preservation.} As seen in Figure~\ref{fig:lfw}, M-DAT keeps more low-frequency words, which verifies our motivation to develop a non-autoregressive multilingual model without the help of knowledge distillation.
In addition to the BLEU scores, this result further provides evidence that our M-DAT has better translation quality than Switch-GLAT.

\textbf{Ablation Study.} Table~\ref{tab:ablation} presents an ablation study on the pivot back-translation (PivotBT) of M-DAT using the IWSLT dataset.
In addition to PivotBT, we consider 3 variants: 1) \emph{M-DAT rand-lang \& w/o pivot}, which
randomly selects an augmented source language but 
does not translate through a pivot language; 2) \emph{ M-DAT src-lang \& w/o pivot}, which directly back-translates the target sentence to the language of the source sentence; and 3) \emph{M-DAT w/o BT}, which does not augment the training data with back-translation. 

\begin{table}[t]
  \centering
  \resizebox{0.9\linewidth}{!}{
  \begin{tabular}{llrrr} \toprule 
  \# &  Model Variant     & Supervised  & Zero-Shot \\ \midrule
  1 & M-DAT (PivotBT)  & \textbf{29.42}      & \bf{19.35}   \\
  2 & ~~  rand-lang \& w/o pivot  &  29.33         &  18.10    \\
  3 & ~~  src-lang \& w/o pivot  & 29.38       & 13.78     \\ 
  4 & ~~  w/o BT  & 28.55    & 13.37     \\ \bottomrule
  \end{tabular}}
  \vspace{-0.2cm}
  \caption{Ablation study on the IWSLT dataset. The results are generated with $n$-gram beam search.}
  \label{tab:ablation}
  \end{table}

In the supervised setting, we observe that back-translation improves the performance (Lines 1--3 vs. Line 4), which is consistent with the findings of previous work \cite{johnson-etal-2017-googles}.  % ,song2022switchglat
Among back-translation methods, \textit{src-lang \& w/o pivot} performs the worst in the zero-shot setting (Line 3). We conjuncture that this is because only applying back-translation to the source language makes the model focus too much on the supervised directions, which degenerates the generalization to the zero-shot setting.
On the other hand, our PivotBT outperforms the direct random back-translation (\textit{rand-lang \& w/o pivot}) and the source-language back-translation  (\textit{src-lang \& w/o pivot}). This confirms that PivotBT provides the model with better-augmented samples for the zero-shot translation.

\section{Conclusion}
In this work, we propose M-DAT to tackle non-autoregressive multilingual machine translation (MNMT). Our approach leverages the recent directed acyclic Transformer so that we do not need the knowledge distillation process, which is particularly inconvenient in the multilingual translation task. Further, we propose a pivot back-translation method to improve the robustness. Our M-DAT achieves state-of-the-art results on supervised and zero-shot settings for non-autoregressive MNMT.

\section{Limitation}
One possible limitation is that our M-DAT obtains slightly lower BLEU scores compared with autoregressive models in the supervised setting. 
However, this is not the drawback of this work, as it is understandable that non-autoregressive models trade quality with more efficiency. Nevertheless, our M-DAT outperforms the previous state-of-the-art NAT approach in both supervised and zero-shot settings, and is easier to be deployed since KD is not required. Our PivotBT, in principle, can also be applied to the training of multilingual autoregressive Transformer (M-AT). However, we do not include M-AT with PivotBT for two reasons: 1) the main focus of this research is on the non-autoregressive Transformer; and 2) the decoding of M-AT is much slower than our M-DAT, which makes the training of M-AT with PivotBT impractical. 

\section*{Acknowledgments}
We would like to thank all reviewers and chairs for their comments.
This research was supported in part by the Natural Science Foundation of China under Grant No. 62376133.
This research was also supported in part by the Natural Sciences and Engineering Research Council of Canada (NSERC) under Grant Nos. RGPIN-2020-04440 and RGPIN-2020-04465, 
the Amii Fellow Program, the Canada CIFAR AI Chair Program, the Alberta Innovates Program, and the Digital Research Alliance of Canada (alliancecan.ca).

\bibliography{custom}
\bibliographystyle{acl_natbib}

\newpage

\clearpage
\appendix

\section{Datasets}
\label{app:datasets}
\begin{table}[t]
\centering
\resizebox{\linewidth}{!}{
\begin{tabular}{lccl}
\toprule
        & Directions &  Languages &Language pairs	\\   \midrule
WMT-EFZ  & 6  & 3	& en$\leftrightarrow$fr, en$\leftrightarrow$zh  \\ 
WMT-EFD  & 6 & 3    & en$\leftrightarrow$fr, en$\leftrightarrow$de  \\ 
WMT-MANY & 10  & 6  & en$\leftrightarrow$fr, en$\leftrightarrow$zh, \\ 
 	     &     &    & en$\leftrightarrow$de, en$\leftrightarrow$ro, \\
 	     &     &    & en$\leftrightarrow$ru                         \\ 
\bottomrule
\end{tabular}
}
\caption{The translation directions of the WMT-EFD, WMT-EFZ, and WMT-MANY datasets.  }
\label{tab:wmt_dataset}
\end{table}

\begin{table*}[t]
\begin{center}
\noindent\resizebox{0.7\linewidth}{!}{\begin{tabular}{|llllllll|} \hline 
\textbf{Supervised Setting}         & AVG   & en-it & en-nl & en-ro & it-en & nl-en & ro-en \\ \hline
M-AT w/ standard layout   & \textbf{30.0} & \textbf{32.7}  & 29.0  & \textbf{24.4}  & \textbf{34.6}  & 29.8  & \textbf{29.5}  \\
M-AT w/ shallow decoder    & \textbf{30.0} & \textbf{32.7}  & 29.0  & \textbf{24.4}  & \textbf{34.6}  & 29.8  & \textbf{29.5}  \\
Residual M-AT & 29.7 & 32.6  & \textbf{29.1}  & 23.9  & 33.8  & 29.6  & 29.3  \\
M-DAT w/ lookahead     & 28.6 & 30.8  & 27.6  & 22.3  & 33.3  & 29.3  & 28.2  \\
~~ w/ $n$-gram beam search   & 29.4 & 32.1  & 28.4  & 23.0    & 34.0  & \textbf{30.0}  & 29.0  \\ \hline  
\end{tabular}}
\resizebox{0.7\linewidth}{!}{\begin{tabular}{|llllllll|}  \hline 
\textbf{Zero-Shot Setting}    & AVG   & it-en & it-ro & nl-it & nl-ro & ro-it & ro-nl \\ \hline
M-AT     & 12.87 & 13.3  & 12.3  & 13.4  & 11.7  & 14.2  & 12.3  \\
M-AT     & 12.87 & 13.3  & 12.3  & 13.4  & 11.7  & 14.2  & 12.3  \\
Residual M-AT & 17.67 & 18.2  & 17.3  & 18.3  & 15.2  & 19.8  & 17.2  \\
M-DAT w/ lookahead   & 18.53 & 20.0    & 17.8  & 19.3  & 15.3  & 20.9  & 17.9  \\
~~ w/ $n$-gram beam search    & \textbf{19.35} & \textbf{20.6}  & \textbf{18.6}  & \textbf{20.3}  & \textbf{16.0}   & \textbf{21.8}  & \textbf{18.8}  \\ \hline 
\end{tabular}}

\end{center}
\caption{BLEU scores on the IWSLT dataset.}
\label{tab:wmt_results_all}
\end{table*}

\textbf{WMT Datasets.}
WMT-EFZ, EFD, and MANY are specifically curated for multilingual machine translation; each is a mix of a few general bilingual machine translation corpora.\footnote{The corpora are obtained from the WMT workshop: \url{https://www.statmt.org/wmt17/}} 
We list the translation directions of the three datasets in Table~\ref{tab:wmt_dataset}. As seen, both WMT-EFZ and WMT-EFD have 3 languages and 6 translation directions, whereas WMT-MANY has 5 languages and 10 directions.  

We strictly followed \newcite{song2022switchglat} for data preparation, and we set the vocabulary size as 85K for WMT-EFD and WMT-EFZ, and 95K for WMT-MANY.

\begin{table*}[ht!]
\begin{center}
\noindent\resizebox{\linewidth}{!}{\begin{tabular}{|l|l|lllll|l|lllll|}
  \hline
        Model &\multirow{5}{*}{\rotatebox{90}{\textbf{WMT-EFD~~~~~~}}}  & AVG   & en-fr & fr-en & en-de & de-en & \multirow{5}{*}{\rotatebox{90}{\textbf{WMT-EFZ~~~~~~}}}& AVG   & en-fr & fr-en & en-zh & zh-en \\ \cline{1-1}\cline{3-7}\cline{9-13}
M-AT w/ standard layout && \textbf{34.57} & \textbf{42.30} & \textbf{37.88} & 25.85 & \textbf{32.23} & &\textbf{31.15} & \textbf{42.14} & \textbf{37.64} &  21.02 & \textbf{23.80} \\
M-AT w/ shallow decoder && 34.19 & 42.19 & 37.39 & \textbf{26.22} & 30.96 &&  30.87  & 42.15 & 37.92 & \textbf{21.44} & 21.98 \\  
Switch-GLAT && 33.34 & 40.81 & 36.00 & 25.27 & 31.29 && 29.76 & 40.54 & 36.48 & 19.47 & 22.55 \\  \cline{1-1}\cline{3-7}\cline{9-13}
M-DAT w/ lookahead && 33.72 & 41.81 & 37.69 & 24.00 & 31.31 & & 30.39 & 41.23 & 36.96 & 20.46 & 22.90 \\
~~ w/ $n$-gram beam search && 33.83 & 41.54 & 37.84 & 24.23 & 31.70 && 30.55 & 41.12 & 37.34 & 20.87 & 22.85 \\
\hline
\end{tabular}}
\resizebox{\linewidth}{!}{\begin{tabular}{|llllllllllll|} \hline 
\textbf{~~~~~~~~~~~~WMT-MANY}   & AVG    & en-de & de-en & en-fr & fr-en & en-ro & ro-en & en-ru & ru-en & en-zh & zh-en \\ \hline
M-AT w/ standard layout & \textbf{30.36} & 24.67 & \textbf{32.16} & 41.67 & \textbf{38.00} & \textbf{32.86} & \textbf{35.65} & \textbf{24.22} & \textbf{30.47} & 20.66 & 23.27 \\
M-AT w/ shallow decoder & 29.19 & \textbf{25.41} & 31.38 & \textbf{41.98} & 37.88 & 30.18 & 34.16 & 21.87 & 26.92 & \textbf{20.90} & 21.27 \\
Switch-GLAT & 28.47 & 24.18 & 30.49 & 39.47 & 36.30 & 31.93 & 32.40 & 24.16 & 28.33 & 16.25 & 21.23 \\ \hline
M-DAT  w/ lookahead & 28.69 & 23.48 & 30.30 & 40.78 & 36.76 & 30.77 & 34.83 & 20.14 & 28.30 & 19.61 & 21.94 \\
~~ w/ $n$-gram beam search & 29.73 & 23.91 & 31.52 & 41.12 & 37.63 & 32.11 & 35.44 & 21.45 & 30.09 & 20.60 & \textbf{23.42} \\
\hline
\end{tabular}}
\end{center}
\caption{BLEU scores on three WMT datasets.}
\label{tab:iwslt_all}
\end{table*}

\textbf{IWSLT Dataset.}
We followed \newcite{disentangle-pos} for the IWSLT dataset, and directly obtained the processed data from their published codebase.\footnote{\url{https://github.com/nlp-dke/NMTGMinor/tree/master/recipes/zero-shot}}
The vocabulary size of IWSLT is 19K. The IWSLT test set contains 3 supervised language pairs: en$\leftrightarrow$ro, en$\leftrightarrow$it, and en$\leftrightarrow$nl. Additionally, it contains 3 zero-shot language pairs: ro$\leftrightarrow$it, ro$\leftrightarrow$nl, and it$\leftrightarrow$nl. The training set for each supervised direction includes 145K samples.

\textbf{Europarl Dataset.}
We further include the Europarl dataset to evaluate the multilingual capability of the proposed method. 
We followed \newcite{disentangle-pos} for data preprocessing. The Europarl test set has 16 supervised directions (containing English) and 56 zero-shot directions (not containing English). The translation directions are detailed in Table~\ref{tab:europarl_per_direction}.

\section{Settings}
\label{app:settings}
\textbf{Evaluation Metrics.} 
We used BLEU \cite{papineni-etal-2002-bleu} to evaluate the translation quality. To ensure a fair comparison with previous work, we applied two BLEU variants. For the WMT datasets, we followed \newcite{song2022switchglat} and applied tokenized BLEU.  For the IWSLT and Europarl datasets, we followed \newcite{disentangle-pos} and adopted SacreBLEU \cite{post-2018-call}. 

We evaluated the latency on a single Tesla V100 GPU with a batch size of 1 to mimic the real-world scenario where the users' requests come one by one. Our evaluation scripts are also available from the released code.

\textbf{WMT and Europarl Datasets.}
We used the Transformer-base configuration \cite{attentionisallyouneed} as the backbone. To train the model, we set the batch size such that it contains 64K tokens, and let the model train for 800K updates. We used the Adam optimizer. The learning rate was warmed up to 5e-4 using 10K updates, and was annealed with the inverse square roots scheduler. The back-translation strength $\lambda$ was set to 0.5. To balance the sizes of different translation directions, we set the upsampling ratio to 1/3. Since the validation set only contains supervised directions for the IWSLT dataset, we further applied the regularization on the encoder representations \cite{encreg} to prevent overfitting to the supervised directions.

Following the setting in most non-autoregressive machine translation studies \cite{gu2018nonautoregressive,gu-kong-2021-fully,song2022switchglat,dslp}, we evaluated both AT and NAT models by averaging the weights of the best 5 checkpoints, which were selected by their BLEU scores on the validation set. 

For the neural architecture, both our M-DAT and the M-AT with the standard layout have 6 encoder layers and 6 decoder layers. The shallow-decoder M-AT has 12 encoder layers and 1 decoder layer. 

\textbf{IWSLT dataset.}
Most of the settings for IWSLT are the same as those for the WMT and Europarl datasets, but we made some adaptations.
Since the IWSLT dataset is smaller, we set the batch size to 32K tokens. Further, we followed \newcite{disentangle-pos}, and set the number of encoders and decoders to 5 for our M-DAT and the M-AT with the standard layout. On the other hand, the M-AT with the shallow-decoder layout M-AT has 10 encoder layers and 1 decoder layer.

\section{Detailed Results}

We list the per-direction BLEU scores of the three WMT datasets in Table~\ref{tab:wmt_results_all}, IWSLT in Table~\ref{tab:iwslt_all}, and Europarl in Table~\ref{tab:europarl_per_direction}. 

As seen, our M-DAT is only slightly outperformed by M-AT \cite{johnson-etal-2017-googles}, and the gap is small. However, our M-DAT outperforms Switch-GLAT in most of the language directions on the WMT datasets.  

Moreover, our proposed M-AT outperforms the strong Residual M-AT model \cite{disentangle-pos} on all zero-shot translation directions of the IWSLT dataset and of most of the zero-shot directions of the Europarl dataset.

\begin{table*}[t]
\centering
\resizebox{0.9\linewidth}{!}{
\begin{tabular}{lccccc} \toprule
Direction        & M-AT w/ standard layout & M-AT w/ shallow decoder & Residual M-AT & M-DAT w/ lookahead & M-DAT w/ $n$-gram beam search \\ \hline
da-en   & \textbf{38.5}                    & 37.6                    & 37.9          & 38.4            & 38.1                            \\
de-en   & \textbf{36.3}                    & 35.4                    & 35.4          & 35.8            & 36                              \\
en-da   & \textbf{36.5}                    & 35.5                    & 36.3          & 34.7            & 35.9                            \\
en-de   & \textbf{28.6}                    & 27.7                    & 27.8          & 26.8            & 28.1                            \\
en-es   & \textbf{43.0}                      & 42.1                    & 42.2          & 41.7            & 42.9                            \\
en-fi   & 21.9                    & 20.5                    & 21.5          & 19.8            & \textbf{22}                              \\
en-fr   & 38.8                    & 38                      & 37.8          & 37.2            & \textbf{39}                              \\
en-it   & \textbf{33.3}                    & 32.7                    & 32.7          & 31.7            & 33.2                            \\
en-nl   & \textbf{30.0}                      & 28.6                    & 29.5          & 28.8            & \textbf{30.0}                              \\
en-pt   & \textbf{39.3}                    & 37.9                    & 38.3          & 37.7            & 38.8                            \\
es-en   & \textbf{43.2}                    & 43.0                      & 42.6          & 42.7            & 42.1                            \\
fi-en   & \textbf{32.7}                    & 31.4                    & 32.3          & 32.6            & \textbf{32.7}                            \\
fr-en   & 38.5                    & 38.2                    & 38.1          & \textbf{38.8}            & 38.6                            \\
it-en   & \textbf{36.6}                    & 36.3                    & 36.3          & 36.1            & 36.0                              \\
nl-en   & \textbf{34.5}                    & 33.3                    & 33.6          & 34.1            & 34.3                            \\
pt-en   & 40.9                    & \textbf{41.0}                      & 40.5          & 40.4            & 40                              \\ 
\textbf{Average} & \textbf{35.79}                   & 34.95                   & 35.18         & 34.83           & 35.48                           \\ \midrule
da-de   & 15.2                    & 9.8                     & 23.85         & 23.5            & \textbf{25.1}                            \\
da-es   & 28.1                    & 14.8                    & 31.08         & 31.7            & \textbf{33.3}                            \\
da-fi   & 13.3                    & 8.1                     & 17.77         & 15.9            & \textbf{18.4}                            \\
da-fr   & 22.0                      & 13.0                    & 28.58         & 28.6            & \textbf{30.8}                            \\
da-it   & 18.4                    & 10.6                    & 26.17         & 25.0            & \textbf{26.8}                           \\
da-nl   & 18.3                    & 8.8                     & 24.24         & 25.1            & \textbf{26.6}                            \\
da-pt   & 24.6                    & 11.0                   & 28.57         & 28.6            & \textbf{30.1}                           \\
de-da   & 8.7                     & 5.4                     & 26.58         & 27.8            & \textbf{29.8}                            \\
de-es   & 21.8                    & 9.8                     & 29.65         & 31.1            & \textbf{32.8}                            \\
de-fi   & 8.9                     & 5.0                       & \textbf{19.02}         & 15.3            & 17.5                            \\
de-fr   & 15.7                    & 8.6                     & 29.2          & 28.5            & \textbf{30.4}                            \\
de-it   & 15.2                    & 6.8                     & \textbf{26.27}         & 24.3            & 26.0                              \\
de-nl   & 17.7                    & 4.9                     & 25.42         & 25.5            & \textbf{27.1}                            \\
de-pt   & 15.4                    & 6.7                     & \textbf{28.83}         & 28.0              & 29.7                            \\
es-da   & 11.2                    & 5.7                     & 29.02         & 29.5            & \textbf{30.7}                            \\
es-de   & 12.0                      & 5.5                     & \textbf{24.85}         & 23.5            & 24.6                            \\
es-fi   & 9.0                       & 5.7                     & \textbf{20.28}         & 17.0              & 19.0                              \\
es-fr   & 26.2                    & 11.1                    & 33.47         & 34.2            & \textbf{35.8}                            \\
es-it   & 19.5                    & 9.0                       & \textbf{31.55}         & 29.6            & 31.0                              \\
es-nl   & 13.7                    & 4.9                     & 25.3          & 25.9            & \textbf{27.1}                            \\
es-pt   & 23.9                    & 9.8                     & 34.24         & 35.5            & \textbf{36.9}                            \\
fi-da   & 9.8                     & 5.8                     & 24.45         & 24.2            & \textbf{26.1}                            \\
fi-de   & 9.2                     & 5.9                     & 20.4          & 19.1            & \textbf{20.9}                            \\
fi-es   & 15.2                    & 9.8                     & 28.74         & 27.6            & \textbf{29.2}                            \\
fi-fr   & 12.9                    & 7.8                     & 25.27         & 25.2            & \textbf{27.0}                              \\
fi-it   & 9.3                     & 6.5                     & \textbf{24.37}         & 21.5            & 23.1                            \\
fi-nl   & 11.3                    & 5.0                       & 20.95         & 21.2            & \textbf{22.9}                            \\
fi-pt   & 12.9                    & 6.4                     & 25.78         & 24.9            & \textbf{26.6}                            \\
fr-da   & 9.8                     & 7.4                     & 25.57         & 26.9            & \textbf{28.6}                            \\
fr-de   & 12.9                    & 6.9                     & 22.08         & 22.2            & \textbf{23.7}                            \\
fr-es   & 26.8                    & 14.2                    & 32.83         & 35.5            & \textbf{36.9}                            \\
fr-fi   & 9.9                     & 7.1                     & 15.56         & 15.4            & \textbf{17.5}                            \\
fr-it   & 19.5                    & 12.5                    & 28.18         & 28.8            & \textbf{30.4}                            \\
fr-nl   & 14.6                    & 6.2                     & 24.33         & 24.6            & \textbf{26.5}                            \\
fr-pt   & 20.8                    & 10.1                    & 30.01         & 32.7            & \textbf{34.0}                              \\
it-da   & 9.4                     & 4.7                     & 24.62         & 25.5            & \textbf{26.8}                            \\
it-de   & 10.5                    & 5.1                     & 21.75         & 20.5            & \textbf{21.9}                           \\
it-es   & 22.2                    & 10.1                    & 31.99         & 33.6            & \textbf{34.6}                            \\
it-fi   & 8.6                     & 4.7                     & \textbf{17.45}         & 14.6            & 16.7                            \\
it-fr   & 22.1                    & 9.9                     & 30.77         & 31.8            & \textbf{33.2}                            \\
it-nl   & 12.5                    & 4.5                     & 22.71         & 23.2            & \textbf{24.8}                            \\
it-pt   & 16.7                    & 7.9                     & 29.8          & 30.4            & \textbf{32.1}                            \\
nl-da   & 14.3                    & 6.8                     & 27.97         & 26.5            & \textbf{28.0}                              \\
nl-de   & 13.8                    & 6.9                     & \textbf{25.1}          & 22.6            & 24.1                            \\
nl-es   & 21.9                    & 13.6                    & 29.78         & 29.7            & \textbf{31.4}                            \\
nl-fi   & 9.0                       & 5.7                     & \textbf{17.19}         & 14.6            & 16.2                            \\
nl-fr   & 19.4                    & 10.7                    & 27.12         & 27.7            & \textbf{29.4}                            \\
nl-it   & 16.4                    & 8.9                     & 24.79         & 23.4            & \textbf{25.1}                            \\
nl-pt   & 18.2                    & 9.2                     & 27.57         & 26.9            & \textbf{28.4}                            \\
pt-da   & 11.9                    & 6.1                     & 26.8          & 28.2            & \textbf{29.3}                            \\
pt-de   & 11.5                    & 6.3                     & \textbf{24.42}         & 22.4            & 23.7                            \\
pt-es   & 28.1                    & 13.2                    & 36.41         & 37.2            & \textbf{38.2}                            \\
pt-fi   & 9.8                     & 5.9                     & 17.26         & 16.4            & \textbf{18.3}                            \\
pt-fr   & 25.0                      & 12.3                    & 33.03         & 34.1            & \textbf{35.0}                              \\
pt-it   & 19.2                    & 9.7                     & 29.96         & 29.5            & \textbf{30.3}                            \\
pt-nl   & 12.8                    & 5.4                     & 24.25         & 25.5            & \textbf{26.4}                            \\ 
\textbf{Average} & 15.84                   & 8.11                    & 26.13         & 25.86           & \textbf{27.44}       \\ \bottomrule
\end{tabular}}
\caption{BLEU scores on the Europarl dataset.}
\label{tab:europarl_per_direction}
\end{table*}

\end{document}